# Development and Validation of a Deep Learning-Based Microsatellite Instability Predictor from Prostate Cancer Whole-Slide Images

Qiyuan Hu[1], Abbas A. Rizvi[1], Geoffery Schau[1], Kshitij Ingale[1], Yoni Muller[1], Rachel Baits[1], Sebastian Pretzer[2], Aicha BenTaieb[2], Abigail Gordhamer[3,4], Roberto Nussenzveig[3,4], Adam Cole[3,4], Matthew O. Leavitt[3,4,5], Rohan P. Joshi[1], Nike Beaubier[1], Martin C. Stumpe[1], Kunal Nagpal[1]*

[1] Tempus Labs, Inc.   [2] Work done while at Tempus Labs, Inc.
[3] PathNet, Inc.   [4] DDx Foundation   [5] Lumea

*corresponding author: kunal.nagpal@tempus.com

## Abstract

**Background:** Microsatellite instability-high (MSI-H) is a tumor agnostic biomarker for immune checkpoint inhibitor therapy. However, MSI status is not routinely tested in prostate cancer, in part due to low prevalence and assay cost. As such, prediction of MSI status from hematoxylin and eosin (H&E) stained whole-slide images (WSIs) could identify prostate cancer patients most likely to benefit from confirmatory testing and becoming eligible for immunotherapy.

**Methods:** Prostate biopsies and surgical resections from de-identified records of consecutive prostate cancer patients referred to our institution were analyzed. Their MSI status was determined by next generation sequencing. Patients before a cutoff date were split into an algorithm development set (n=4015, MSI-H 1.8%) and a paired validation set (n=173, MSI-H 19.7%) that consisted of two serial sections from each sample, one stained and scanned internally and the other at an external site. Patients after the cutoff date formed the temporal validation set (n=1350, MSI-H 2.3%). Attention-based multiple instance learning models were trained to predict MSI-H from H&E WSIs.

**Findings:** The MSI-H predictor achieved area under the receiver operating characteristic curve values of 0.78 (95% CI [0.69-0.86]), 0.72 (95% CI [0.63-0.81]), and 0.72 (95% CI [0.62-0.82]) on the internally prepared, externally prepared, and temporal validation sets, respectively. While MSI-H status is significantly correlated with Gleason score, the model remained predictive within each Gleason score subgroup.

**Interpretation:** We developed and validated an AI-based MSI-H diagnostic model on a large real-world cohort of routine H&E slides, which effectively generalized to externally stained and scanned samples and a temporally independent validation cohort. This algorithm has the potential to direct prostate cancer patients toward immunotherapy and to identify MSI-H cases secondary to Lynch syndrome.

# Introduction

Prostate cancer is the second most common cancer in the United States, with approximately one in eight men receiving a prostate cancer diagnosis in their lifetime, and represents the fifth most common cause of cancer mortality.[1] Despite an increasing number of targeted and immunotherapy treatment options in cancer overall,[2] use of these modalities has lagged in prostate cancer. Comprehensive next-generation sequencing (NGS) testing is not standard of care and testing rates are low. Mismatch repair (MMR) protein immunohistochemistry (IHC) is also not standard of care because the low prevalence of dMMR prostate cancer makes universal testing cost prohibitive, compared with colorectal cancer, for example, where the higher prevalence has allowed testing to become standard.[3]

Microsatellite instability-high (MSI-H) is a biomarker caused by a deficiency in DNA mismatch repair and is associated with response to immune checkpoint inhibitor therapy. While there is not currently a prostate cancer specific approval, pembrolizumab, a PD-1 inhibitor, has a tumor-agnostic approval for use in MSI-H or dMMR solid tumors. In prostate cancer, MSI-H is uncommon and has been reported at only 2-3% prevalence.[4,5] However, overall response rates to immune checkpoint inhibitors of 25-60%, including durable responses, have been reported across several small studies in this subgroup of patients.[4,6,7] Biomarker-unselected prostate cancer populations have shown limited benefit from immune checkpoint inhibitors, highlighting the critical importance of biomarker testing for MSI-H or dMMR for prostate cancer immunotherapy to enrich for responders despite its low prevalence.[8–10]

While MSI testing is not routinely performed for prostate cancer patients, prostate cancer diagnosis nearly always involves a tissue biopsy with hematoxylin and eosin (H&E)-stained slides and residual formalin-fixed paraffin-embedded (FFPE) tissue, which can be used for NGS or IHC stains. The H&E stained slides are increasingly being digitized as whole slide images (WSIs) to assist pathology workflows and for archival purposes. Therefore, predicting MSI status from H&E WSIs is potentially impactful for identifying patients who are likely MSI-H and may benefit from confirmatory testing and immunotherapy and/or Lynch syndrome testing.

The application of machine learning on WSIs has been studied for predicting MSI-H in colorectal and gastric cancers.[11–13] However, MSI prediction in prostate cancer has been less well studied, with lower prevalence and lack of testing in the standard of care posing challenges to collecting sufficient positive samples. This dearth of testing in standard care also creates an unmet need to identify MSI-H/dMMR tumors, and H&E-based machine learning models could assist in narrowing down the population to be tested so that it becomes feasible to do so. Moreover, the generalizability of histopathology machine learning algorithms across multi-site staining and scanning characteristics remains a significant challenge, and validating algorithm performance across external pre-analytic characteristics remains important for algorithm utility.[14]

In this study, we developed a machine learning model to predict MSI-H from a large, real-world prostate cancer cohort containing WSIs, clinical data, molecular testing results, and IHC assay results. We directly validated the generalizability of the predictor to stain and scanner characteristics by evaluating performance on an externally prepared dataset composed of a serial section of each slide from the internal validation set but stained at a different site and scanned using a different scanner model. Finally, we validated the model's generalizability to a temporally independent internal validation cohort. The predictor demonstrated high effectiveness in identifying MSI-H from WSIs and has the potential to identify prostate cancer patients most likely to benefit from confirmatory testing for evaluation of their immunotherapy eligibility and Lynch syndrome testing.

# Methods

## Study design and participants

In this retrospective, diagnostic study, we sampled de-identified records of consecutive prostate cancer patients who were sequenced at Tempus Labs (Chicago, IL, USA) from October 2017 to February 2023 and whose WSIs of prostate biopsies or surgical resections were available. Each case included clinical characteristics, molecular profiles, and digitized WSIs. About 25% of cases have MMR protein IHC. All samples were digitized with either a Philips UFS scanner or an Aperio GT 450 scanner, and 37% of samples were stained at external laboratories. MSI-H/microsatellite-stable (MSS) status was determined using DNA NGS. Samples with equivocal or undetermined MSI status (n=97), samples that failed quality control by pathologists (n=88), and samples with Gleason scores of less than 7 (n=48) were excluded (Fig. 1). Reasons for equivocal or undetermined MSI status predominantly result from insufficient tumor purity for the MSI call in NGS, but may also include sequencing of insufficient depth over enough of the assayed microsatellite loci to prevent the prediction from reaching statistical significance, or in rare cases (n=2) multiple tests for the same patient returning conflicting results.

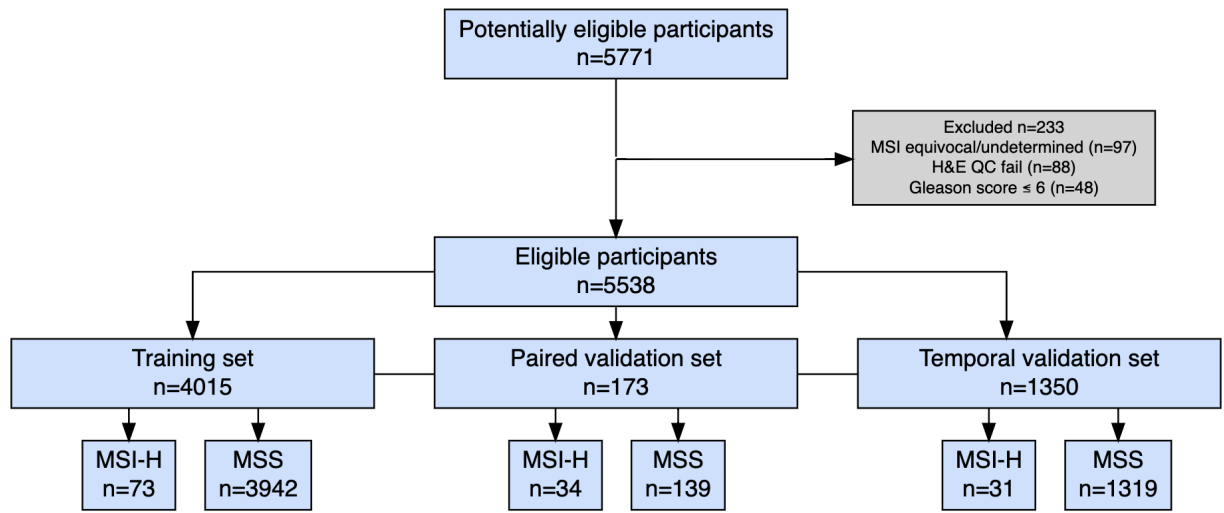

Figure 1. STARD diagram of the study.

A cutoff sequencing date in July 2022 was selected to split the cohort into two temporally independent subsets. Patients sequenced before a cutoff date (n=4188, MSI-H 2.6%), formed the training set and a paired validation set. The training set (n=4015, MSI-H 1.8%) was used for model development and the paired set (n=173, MSI-H 19.7%) to directly evaluate stain and scanner generalizability. The paired validation set was composed of two serial sections from each sample, one of which was stained and scanned internally and another stained and scanned at an external site, TruCore Pathology (Little Rock, AR), using an Aperio AT2 scanner. This set was constructed by randomly sampling 36 MSI-H cases where an unstained serial section was available for study use, and correspondingly sampling 144 MSS cases with matched Gleason score and procedure type distribution as the selected MSI-H samples prior to quality control exclusions. Patients sequenced after the cutoff date formed the temporal validation set (n=1350, MSI-H 2.3%), which was used to evaluate model generalizability on temporally independent data. The design of the data cohorts is illustrated in Fig. 2(a) and 2(b).

## Model development

Tissue and marker regions were first identified on WSIs using a previously developed U-Net model.[15] Subsequently, tiles of size 256x256 pixels at 20x magnifications were generated from WSIs. Tiles predicted to not contain tissue or to contain markers were excluded. International Color Consortium (ICC) profile transformations[16] were applied to correct the color discrepancies between Leica GT450, Leica AT2, and Philips UFS scanners.

An attention-based multiple instance learning model similar to Ilse et al. was trained using these images,[17] as illustrated in Fig. 2(c). The model accepts a "bag of tiles" as input. An ImageNet pre-trained ResNet18 model was used as a feature extraction module for each tile,[18] while the attention module was used to identify tiles with high diagnostic relevance and aggregate information from all tiles in the bag to make a slide-level prediction. The entire model was trained end-to-end using Adam optimizer and weighted cross-entropy loss where weights were assigned according to the class prevalence.[19] In each epoch during training, 200 tiles were randomly sampled to form a bag. The effective batch size was 32 during training, split across four NVIDIA A100 GPUs. At inference time, the bag size was increased to 1600 with a batch size of 1 on one NVIDIA A100 GPU. Tile sampling was performed without replacement, and if a slide had fewer tiles than the bag size, all tiles in the slide were used. Tiles were normalized with the mean and standard deviation of a reference set of H&E images. For data augmentation during training, tiles were randomly cropped to 224x224, randomly rotated by multiples of 90 degrees, randomly flipped, and randomly applied with color jittering.

Five-fold cross-validation within the training set was used to perform hyperparameter tuning to select the learning rate, weight decay, dropout rate, patience and minimum delta for early stopping, input image magnification, and color augmentation parameters (see Supplementary Materials for detailed information). Data splitting for creating cross-validation folds was done such that MSI status and potential confounding variables, such as scanner type, procedure type, and Gleason score were represented equally in each fold. Once the final hyperparameters were selected, the MSI-H predictor was composed by averaging the predictions across the five models trained via cross-validation using the selected hyperparameters. This predictor was finally evaluated on three validation sets: the paired validation set with enriched MSI-H prevalence composed of internally and externally stained and scanned serial sections for each sample, as well as on the temporal validation set to evaluate temporal generalizability.

## Evaluation

The area under receiver operating characteristic curve (AUC) was used as the main metric to evaluate classification performance. Sensitivity, specificity, positive predictive value (PPV), and negative predictive value (NPV) were also reported to assess model performance at various target sensitivity levels. The Pearson correlation coefficient, $R$, was used to evaluate the

(a) Data cohorts

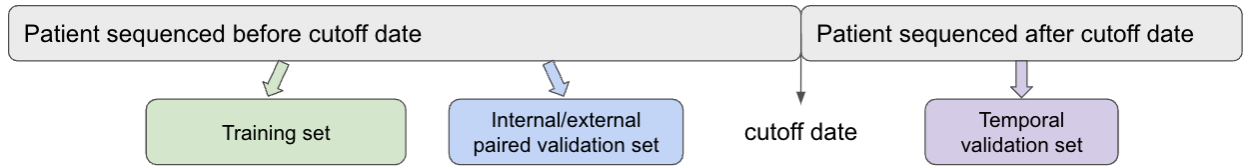

(b) Paired validation set

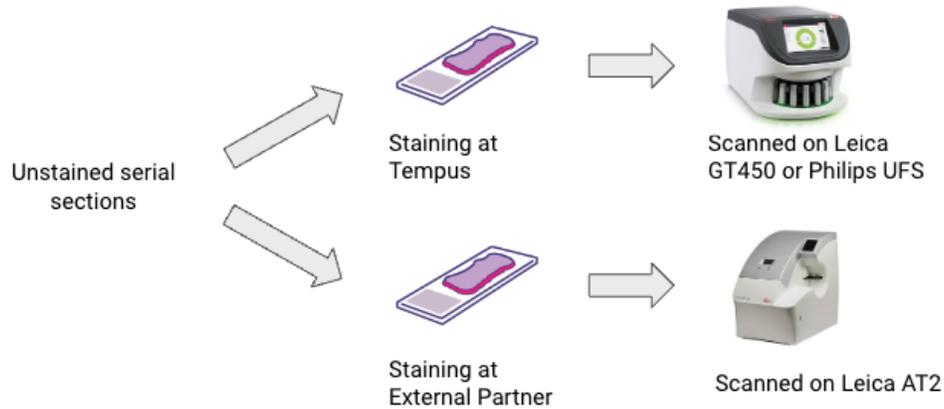

(c) Model prediction pipeline

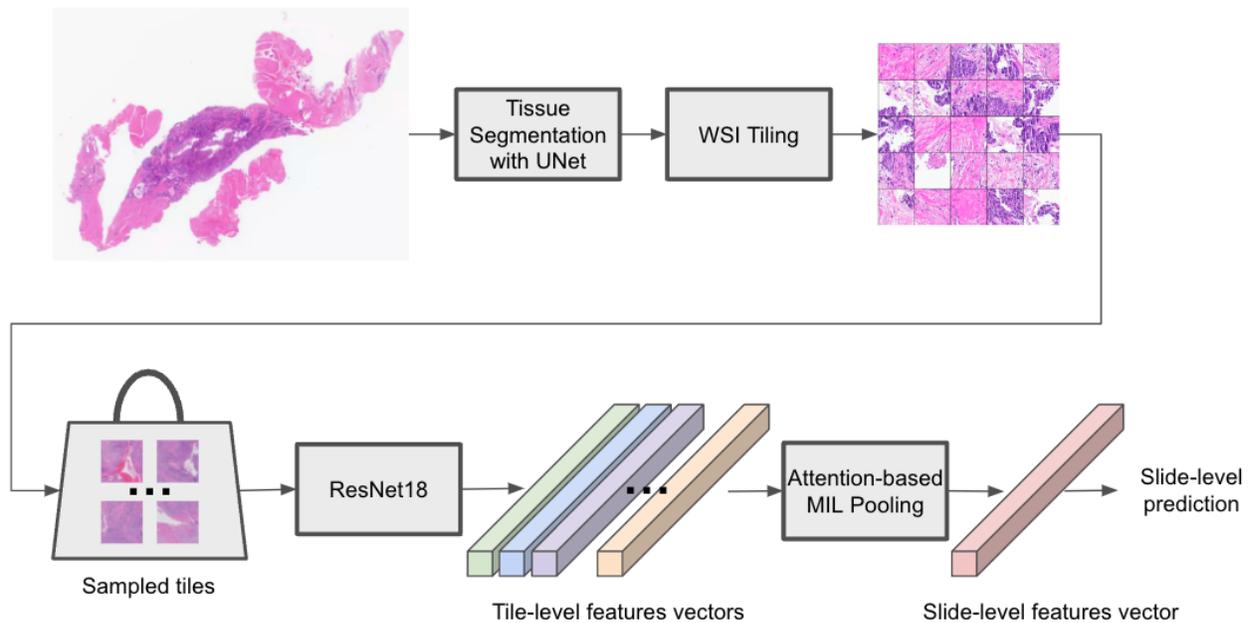

Figure 2. Schematic representation of (a) data cohorts, (b) paired validation set, and (c) model prediction pipeline. Scanner images in (b) attributed to tissuepathology.com and myco-instrumentation.com.[29,30]

correlation between predictions on internally and externally stained and scanned images in the paired validation set. The 95% confidence intervals (CIs) of all metrics were calculated by bootstrapping the prediction scores with 1000 bootstrap samples.

To assess the robustness of model performance across Gleason score and procedure type, subgroup analyses were performed on the pooled internal validation set, which combined the temporal validation set and the internally stained and scanned slides from the paired validation set.

Additional analyses were performed on the pooled internal validation set to assess the robustness of model performance in two challenging subgroups: specimens with small tissue area or low tumor purity. Samples with tissue area or tumor purity in the lowest quartile in the validation set formed these subgroups, corresponding to ≤ 9.35 mm$^2$ tissue area and ≤ 50% tumor purity. A simulation experiment was also performed to establish the model's limit of detection. We performed model inference with different bag sizes, randomly sampling 3, 6, 12, 25, 50, 100, 200, 400, and 800 tiles from each slide in the validation set. Tiles were oversampled if a slide had fewer than the required number of tiles. The experiment was repeated 10 times with different random seeds. The five-fold cross-validation ensemble AUC was calculated for each bag size.

We evaluated how the amount of training data would affect the algorithm performance in a data titration experiment. The training set was consecutively sub-sampled without replacement to 80%, 60%, 40%, and 20% of the original size, stratified by MSI status. A model was trained on each of these subsets using the same hyperparameters and configurations as the original model developed on the full training set, and model performance was analyzed to investigate the impact of sample size available for model development.

Finally, attention heatmaps as well as high- and low-attention tiles from samples in the validation sets were visualized to inspect regions that the model deemed important in making slide-level predictions. Pathologists reviewed randomly sampled high- and low-attention tiles from MSI-H and MSS slides as a sanity check for the model behavior and to identify prominent features.

## Statistical analysis

For analyzing variable correlations with MSI status in the cohort characteristics tables, the Wilcoxon rank-sum test was used for continuous variables, the Pearson's Chi-square test was used for categorical variables when no expected cell count was less than five, and the Fisher test was used for categorical variables when any expected cell count was less than five. The Mann-Whitney U test was used to compare the prediction score distributions between MSI-H and MSS samples in the subgroup analysis. A p < 0.05 was considered to indicate a statistically significant difference. All statistical analyses were done using R 4.2.3 (package: gtsummary 1.7.1) and Python version 3.7.12 (package: statannotations version 0.5.0).[20,21]

# Results

## Cohort characteristics

Table 1 and Supplementary Table 1 present the patient characteristics of the cohorts. A multivariate logistic regression model that predicts MSI status based on the clinical and demographic variables of the cohort shows that Gleason score, sample collection date, and tumor mutational burden (TMB) have statistically significant coefficients (Supplementary Table 2). Other variables that showed significant univariate correlation with MSI status in the cohort tables did not remain significant in multivariate analysis. Higher Gleason scores are associated with greater MSI-H prevalence, ranging from 0.6% amongst Gleason 7 cases to 8.5% amongst Gleason 10 cases. No other significant correlations were found between MSI status and clinical or demographic variables. Table 2 characterizes the MMR results where IHC stains were also available for MSI-H cases, as well as counts of detected somatic and germline MMR gene mutations. MSH2/MSH6 absence was the most common abnormal staining pattern, occurring in 32/38 (84.2%) of cases. Four cases were MLH1/PMS2 absence (10.5%), one case was PMS2-only absence (2.6%), and one case had no MMR protein loss detected (2.6%) but NGS detected an MSH6 mis-sense mutation, E1193K, which has previously been determined to impair heterodimerization with MSH2 and resulting MMR capability.[22]

## Model performance

An attention-based multiple instance learning network was trained on tiles randomly sampled from H&E WSI tissue regions to predict MSI-H. The MSI-H predictor achieved AUC values of 0.78 (95% CI [0.69-0.86]), 0.72 (95% CI [0.63-0.81]), and 0.72 (95% CI [0.62-0.82]) on internally stained and scanned, externally stained and scanned, and temporal validation sets, respectively (Fig. 3). Significant correlation between prediction scores on the paired internally and externally stained and scanned serial sections was observed (R = 0.85, 95% CI [0.77, 0.91], Supplementary Fig. 1). At an example operating point of 50% sensitivity, the MSI-H predictor had a specificity of 86.8% (95% CI [59.9%, 95.7%]), a PPV of 7.9% (95% CI [2.7%, 23.2%]), and an NPV of 98.6% (95% CI [97.9%, 99.1%]) on the temporal validation set. The PPV is notably higher than the underlying MSI-H prevalence of 2.3%. A review of the high-attention tiles suggests the predictor focuses on dense tumor regions in making its determination, while its low-attention tiles largely comprise tiles with stroma and whitespace (Supplementary Fig. 2).

We assessed performance within subgroups on a pooled validation set combining the internally stained and scanned images in the paired validation and the temporal validation sets (Fig. 4). The ROC curves and the violin plots of prediction scores show that the model remained predictive of MSI-H status within each Gleason score and procedure type subgroup. AUC trended higher in the Gleason scores 7-8 subgroup (AUC=0.80, 95% CI [0.66, 0.94]). In the Gleason scores 9-10 subgroup, where MSI-H prevalence is the highest, patients are classified as high-risk, and the need for therapy is often significant, the AUC was also encouraging (AUC=0.72, 95% CI [0.64, 0.81]), and the distributions of prediction scores for MSI-H and MSS patients were significantly different. Performance within surgical resections trended higher than within biopsies (AUC=0.86, 95% CI [0.77, 0.95] vs. AUC=0.73, 95% CI [0.65, 0.80]), and the distributions of prediction scores for MSI-H and MSS patients were significantly different in both subgroups, potentially owing to larger tissue context and reduced frequency of biopsy-related artifacts. Subgroup analysis within each validation set shows qualitatively similar trends but did not have adequate statistical power to assess significance in several subgroups owing to smaller sample sizes (Supplementary Fig. 3).

Table 1. Patient characteristics in data cohorts

Table 1. Patient Characteristics in data cohorts

| Variable | Overall | | | Training set | | Paired validation set | | Temporal validation set | |
|---|---|---|---|---|---|---|---|---|---|
|  | MSI-H, N = 138[1] | MSS, N = 5,400[1] | p-value[2] | MSI-H, N = 73[1] | MSS, N = 3,942[1] | MSI-H, N = 34[1] | MSS, N = 139[1] | MSI-H, N = 31[1] | MSS, N = 1,319[1] |
| **Age at Collection Date** | 71 (65, 76) | 66 (60, 73) | <0.001 | 71 (64, 76) | 66 (60, 72) | 71 (65, 76) | 67 (61, 73) | 69 (63, 77) | 68 (61, 74) |
| Unknown | 14 | 922 |  | 6 | 532 | 1 | 17 | 7 | 373 |
| **Race** |  |  | 0.5 |  |  |  |  |  |  |
| Asian | 3 (4.2%) | 67 (2.8%) |  | 1 (2.9%) | 56 (3.0%) | 1 (4.8%) | 1 (1.4%) | 1 (6.7%) | 10 (2.1%) |
| Black or African American | 11 (15%) | 461 (19%) |  | 2 (5.7%) | 348 (19%) | 4 (19%) | 11 (15%) | 5 (33%) | 102 (21%) |
| White | 57 (80%) | 1,896 (78%) |  | 32 (91%) | 1,462 (78%) | 16 (76%) | 62 (84%) | 9 (60%) | 372 (77%) |
| Unknown | 67 | 2,976 |  | 38 | 2,076 | 13 | 65 | 16 | 835 |
| **Histology** |  |  | 0.2 |  |  |  |  |  |  |
| Adenocarcinoma | 135 (98%) | 5,317 (98%) |  | 71 (97%) | 3,879 (98%) | 33 (97%) | 134 (96%) | 31 (100%) | 1,304 (99%) |
| Carcinoma | 2 (1.4%) | 17 (0.3%) |  | 1 (1.4%) | 15 (0.4%) | 1 (2.9%) | 1 (0.7%) | 0 (0%) | 1 (<0.1%) |
| Neuroendocrine | 1 (0.7%) | 50 (0.9%) |  | 1 (1.4%) | 36 (0.9%) | 0 (0%) | 2 (1.4%) | 0 (0%) | 12 (0.9%) |
| Sarcoma | 0 (0%) | 3 (<0.1%) |  | 0 (0%) | 3 (<0.1%) |  |  |  |  |
| Small cell carcinoma | 0 (0%) | 13 (0.2%) |  | 0 (0%) | 9 (0.2%) | 0 (0%) | 2 (1.4%) | 0 (0%) | 2 (0.2%) |
| **Total Gleason** |  |  | <0.001 |  |  |  |  |  |  |
| 7 | 5 (4.3%) | 870 (21%) |  | 3 (4.8%) | 691 (22%) | 0 (0%) | 9 (8.2%) | 2 (8.7%) | 170 (19%) |
| 8 | 18 (15%) | 825 (20%) |  | 9 (15%) | 617 (20%) | 3 (9.4%) | 11 (10%) | 6 (26%) | 197 (22%) |
| 9 | 61 (52%) | 2,078 (50%) |  | 33 (53%) | 1,602 (51%) | 17 (53%) | 50 (45%) | 11 (48%) | 426 (49%) |
| 10 | 33 (28%) | 368 (8.9%) |  | 17 (27%) | 245 (7.8%) | 12 (38%) | 40 (36%) | 4 (17%) | 83 (9.5%) |
| Unknown | 21 | 1,259 |  | 11 | 787 | 2 | 29 | 8 | 443 |
| **Procedure Type** |  |  | 0.5 |  |  |  |  |  |  |
| Ambiguous Biopsy | 18 (13%) | 578 (11%) |  | 11 (15%) | 431 (12%) | 3 (9.1%) | 12 (8.9%) | 4 (13%) | 135 (10%) |
| Core needle biopsy | 80 (59%) | 2,910 (56%) |  | 42 (58%) | 2,009 (54%) | 18 (55%) | 77 (57%) | 20 (65%) | 824 (63%) |
| Resection + excisional | 38 (28%) | 1,691 (33%) |  | 19 (26%) | 1,293 (35%) | 12 (36%) | 46 (34%) | 7 (23%) | 352 (27%) |
| Unknown | 2 | 221 |  | 1 | 209 | 1 | 4 | 0 | 8 |

[1] Median (IQR); n (%)

[2] Wilcoxon rank sum test; Fisher's exact test; Pearson's Chi-squared test

Table 2. Distribution of mismatch repair (a) immunohistochemical (IHC) stain findings and (b) gene mutations for MSI-H prostate cancer cases. For the one MSI-H case where all MMR staining patterns were present, NGS detected an MSH6 mis-sense mutation (E1193K).

| IHC Staining Pattern | Number of samples |
|---|---|
| MSH2/MSH6 Loss | 32 |
| MLH1/PMS2 Loss | 4 |
| MSH6 Loss Only | 0 |
| PMS2 Loss Only | 1 |
| All Present | 1 |
| MMR IHCs not available | 100 |

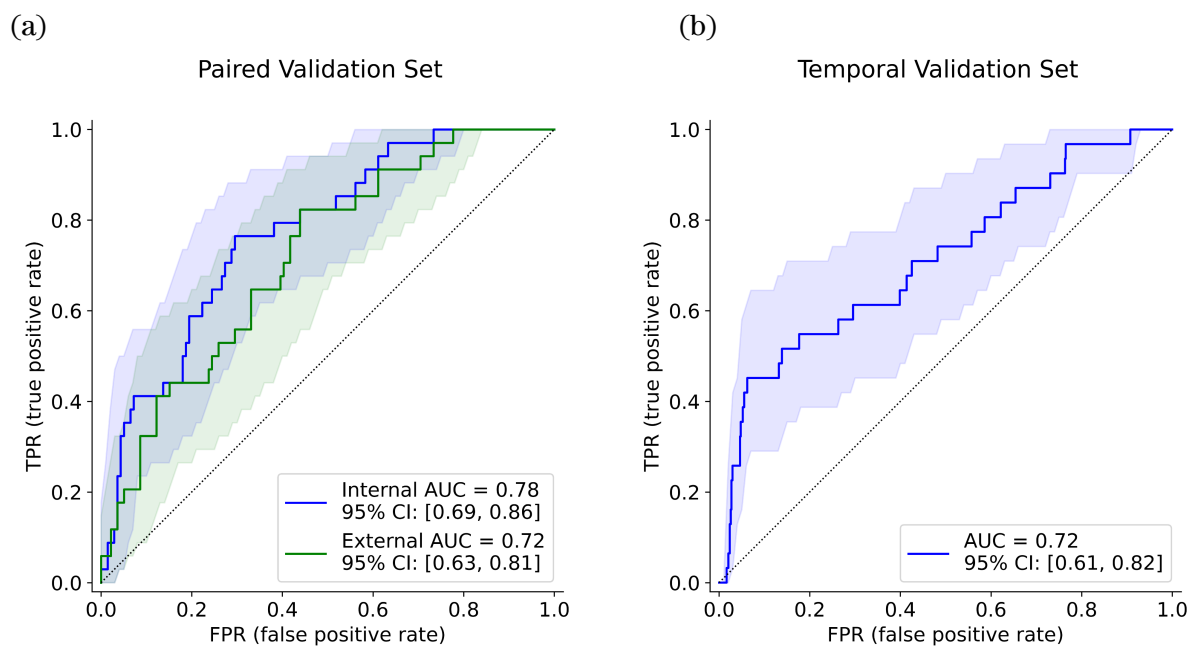

| Sensitivity | Specificity | PPV | NPV |
| --- | --- | --- | --- |
| 50% | 86.8% [59.9%, 95.7%] | 7.9% [2.7%, 23.2%] | 98.6% [97.9%, 99.1%] |
| 70% | 57.5% [39.0%, 85.3%] | 3.8% [2.3%, 10.2%] | 98.8% [97.9%, 99.4%] |
| 90% | 26.9% [22.0%, 53.1%] | 2.8% [2.0%, 4.5%] | 99.2% [98.6%, 99.7%] |

Figure 3. Receiver operating characteristic (ROC) curve for the MSI-H predictor on the paired validation set and the temporal validation set, and tables of metrics and their 95% confidence intervals at various target sensitivities on the temporal validation set.

Additional subgroup analyses showed that the algorithm performance remained robust on small specimens with tissue area in the lowest quartile (AUC=0.76, 95% CI [0.61, 0.92]) and trended slightly lower on samples with tumor purity in the lowest quartile (AUC=0.71, 95% CI [0.61, 0.83]) (Supplementary Fig. 4(a)). The simulation experiment showed that the model performance remained robust down to bag sizes of 50-100 tiles, corresponding to 0.6-1.3 mm$^2$ of sampled tissue area, which is the 0.01 percentile in our dataset and is much smaller than a core needle biopsy (Supplementary Fig. 4(b)).

Finally, the data titration experiment showed that model performance on the validation sets increased as a larger fraction of training data was used, and the model performance may yet improve with additional training data (Supplementary Fig. 5).

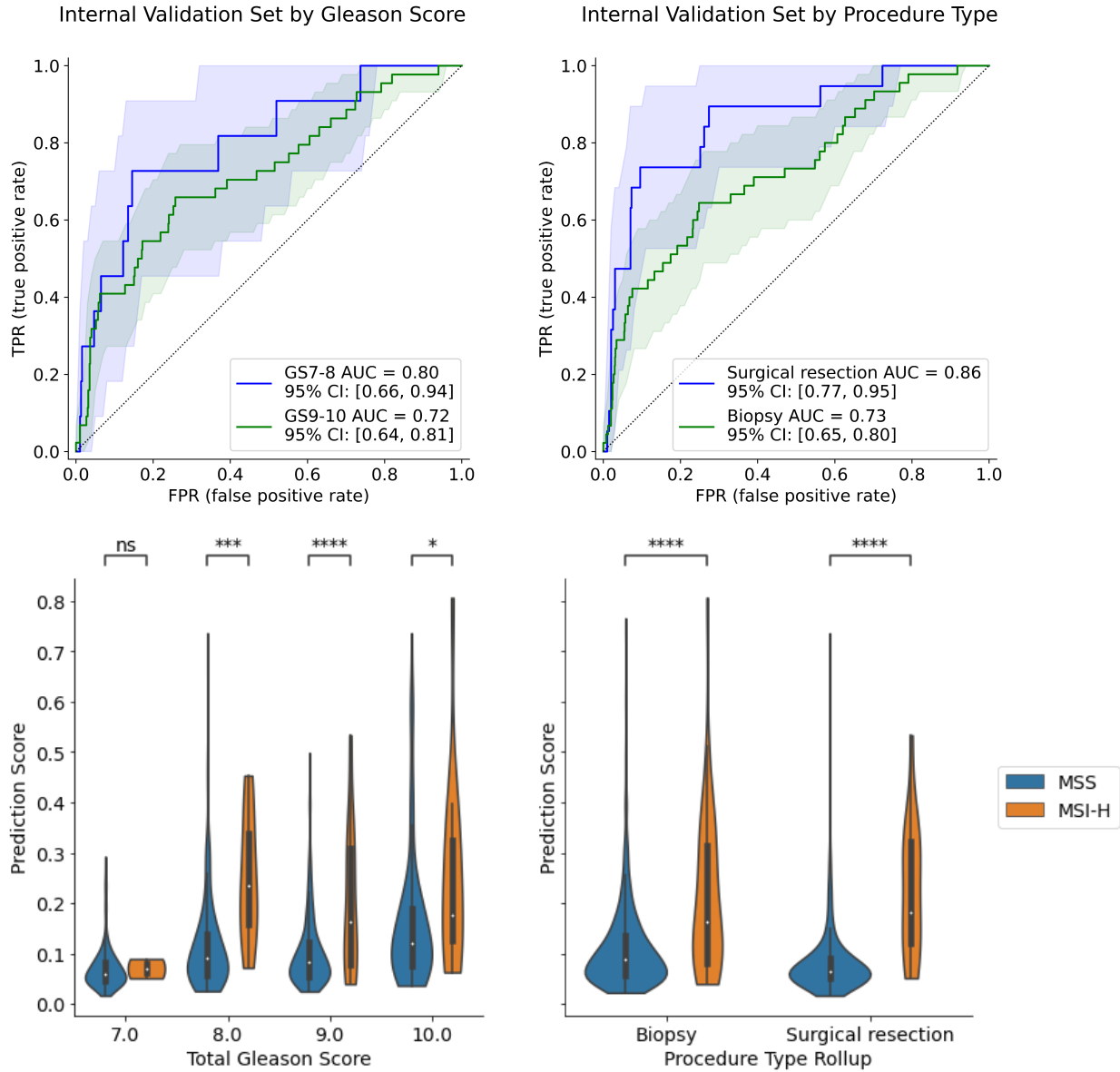

Figure 4. Receiver operating characteristic (ROC) curves and violin plots of prediction scores for various clinical subgroups in the pooled validation set that combines the internal scans of the paired validation set and the temporal validation set. P-value annotation legend: ns: p ≤ 1, *: 0.01 < p ≤ 0.05, **: 0.001 < p ≤ 0.01, ***: 0.0001 < p ≤ 0.001, ****: p ≤ 0.0001.

# Discussion

In this study, we developed a deep learning predictor of MSI status using a large, real-world cohort of H&E whole slide images and corresponding molecular testing results and evaluated its generalizability to externally stained and scanned slides and to a temporally independent validation cohort. The predictor achieved high performance for a screening algorithm and demonstrated significant discriminative ability on both the externally stained and scanned images and the temporal validation set. Given the predictor's effectiveness and generalizability, the ubiquity and increasing digitization of H&E slides in prostate cancer diagnoses, and the lack of routine testing for MSI in prostate cancer, we anticipate that our algorithm could be used to direct testing and find patients eligible for targeted therapies who otherwise may have been missed.

For patients determined to be MSI-H via confirmatory testing, the clinical implications are significant, including potential eligibility to receive pembrolizumab, which has a tumor-agnostic indication in MSI/MMRd tumors and reported response rates of 25-60%.[4,6–8] Other immunotherapies may also be effective, with evidence of encouraging response rates to nivolumab, a PD-1 inhibitor in a Phase II clinical trial.[23] These findings show that our MSI predictor is potentially impactful on patient outcomes.

Furthermore, for cases where MMR and MSI were both evaluated in this study, loss of MSH2 and MSH6 expression were the most common reasons for MMR deficiency. Given the association of concurrent MSH2/MSH6 loss with MSH2 gene mutations, these patients may have a higher likelihood of Lynch syndrome.[24] Lynch syndrome, caused by a defective mismatch repair due to the presence of a germline mutation in an MMR gene, predisposes carriers to significantly elevated lifetime cancer risk,[24] and more stringent screening protocols for these patients are warranted.[25] Therefore, the clinical value of MSI-H detection in these cases may extend beyond treatment selections in the patients themselves, but also to directing germline testing and surveillance protocols for their families for heritable cancer predisposition syndrome.

Subgroup analyses showed that the model remained predictive within Gleason score subgroups, including scores of 9-10 where the impact of this algorithm may be the greatest. Patients exhibiting Gleason score 9-10 (Grade Group 5) have significantly worse prognosis than other prostate cancer patients,[26] are minimally considered stage IIIC independent of metastatic status,[27] and correspondingly tend to receive aggressive treatment including hormonal and radiation therapy. The MSI-H prevalence is also greatest amongst these patients in our study, and similar associations have been noted in other studies.[28] While MSI-H has been associated with favorable prognosis in other cancer types, the prognostic significance of MSI in high Gleason score prostate cancer is not yet fully understood, and treatment selection for these patients remains a significant need. Given the greater prevalence and significant clinical need for treatment in high-grade prostate cancers, we anticipate the predictor's utility and urgency of MSI-H confirmatory testing may be the greatest in this subgroup.

Compared with the model performance on the internally stained and scanned slides, the performance dropped slightly on the externally stained and scanned slides in the paired validation set, which could be attributed to differences in staining and scanning protocols, a well-known challenge for the generalizability of deep learning algorithms in digital pathology. We employed robustness measures, such as ICC profile transformation and color augmentation during model training, to reduce domain shift and increase model generalizability. Additionally, although all patients were sequenced at Tempus, 37% of H&E slides in our cohort were prepared externally at other laboratories using different staining protocols and scanners, mimicking diverse data in a multi-institutional study and adding confidence to the generalizability of our model. The model performance is also slightly lower on the temporally independent validation set than the internal slides in the paired validation set, which could be attributed to temporal drift in data distribution, evidenced by the significant association between sample collection year and MSI status, and in the color space of H&E slides as staining and scanning techniques evolved over time.

There are a few limitations to our work. First, the performance and generalizability that our model achieved were constrained by the limited number of MSI-H cases in our cohort and may benefit from additional data. Next, performance was evaluated on one slide per case while several slides are typically produced for each case during typical pathology workflows. Future work is needed to study the selection of optimal slides for the predictor to analyze for a given case. While the model demonstrated potential utility on small biopsy specimens, subgroup analyses showed trends toward algorithm performance degradation at low tumor purities, suggesting the algorithm may be most applicable on specimens with high tumor purity. Finally, although we constructed a paired validation set to evaluate model generalizability to externally stained and scanned slides, our real-world dataset has inherent biases stemming from the retrospective inclusion of only patients who underwent sequencing and the utilization of single-institution sequencing results. As such, a prospective, multi-institution evaluation of the MSI-H predictor is warranted prior to the algorithm's use in clinical practice.

# Acknowledgments

We thank Robert Montgomery for his management support and support of machine learning infrastructure tooling that enabled this work. We thank Adam Hockenberry and Matthew Kase for reviewing and editing the manuscript. We thank Sun Hae Hong and Josh Och for their constructive feedback on the manuscript. We thank Abigail Leavitt for helping with slide digitization.

# Supplementary Materials

## Hyperparameter tuning

Five-fold cross-validation within the training set was used to perform hyperparameter tuning to select for learning rate, weight decay, dropout rate, patience and minimum delta for early stopping, input image magnification, and color augmentation parameters. Learning rate, weight decay, and dropout rate were combined into one grid search, while other hyperparameters were swept individually on a grid search. The ranges of the search were image magnification: 5x, 10x, 20x; learning rate: 1e-7~1e-4; weight decay: 1e-5~1e-2; dropout rate: 0.1~0.5; early stopping patience: 5~20; early stopping minimum delta: 0~0.001; color augmentation: spans the entire range of color jittering. The set of hyperparameters that achieved the highest ensemble validation AUC during cross-validation was selected as optimal, which was image magnification 20x, learning rate 1e-6, weight decay 1e-4, dropout rate 0.1, early stopping patience 10, early stopping minimum delta 0.00025, slide-level brightness augmentation 0.25, slide-level contrast augmentation 0.5, slide-level saturation augmentation 0.25, and slide-level hue augmentation 0.04.

## Model generalizability

Our unique paired serial section validation set allowed us to measure the impact of methods on model robustness, as measured by the correlation between predictions between the internally and externally stained and scanned serial sections, in addition to overall model predictive performance. Our model showed a correlation of 0.73 (95% CI [0.60, 0.83]) (Supplementary Fig. 1).

We conducted exploratory analyses to assess the impact of several methods on our MSI-H predictor. The following methods resulted in notable effects (Supplementary Fig. 6). First, compared with tile-level color augmentation, slide-level color augmentation showed improvements in model generalizability, with a significant increase in R of 0.13 (95% CI [0.05, 0.26]) between predictions on the internally and externally stained and scanned slides in the paired validation set. On the other hand, contrary to what some other studies have suggested, self-supervised pre-training reduced model performance and generalizability, with a decrease in AUC of -0.02 (95% CI [-0.06, 0.01]) and -0.09 (95% CI [-0.19, -0.00]) on the internally and externally stained and scanned slides, respectively, and a significant change in R of -0.35 (95% CI [-0.47, -0.18]). The 95% CIs of the difference in metrics were calculated by bootstrapping the paired prediction scores from two models with 1000 bootstrap samples.

## Supplementary Table 1. Patient clinical and molecular features

| Variable | Overall | | | Training set | | Paired validation set | | Temporal validation set | |
|---|---|---|---|---|---|---|---|---|---|
| | MSI-H, N = 138[1] | MSS, N = 5,400[1] | p-value[2] | MSI-H, N = 73[1] | MSS, N = 3,942[1] | MSI-H, N = 34[1] | MSS, N = 139[1] | MSI-H, N = 31[1] | MSS, N = 1,319[1] |
| **Stage** | | | 0.059 | | | | | | |
| Stage 1 | 2 (2.3%) | 267 (8.4%) | | 0 (0%) | 222 (9.0%) | 2 (8.0%) | 5 (5.9%) | 0 (0%) | 40 (6.2%) |
| Stage 2 | 2 (2.3%) | 73 (2.3%) | | 1 (2.3%) | 63 (2.6%) | 0 (0%) | 1 (1.2%) | 1 (5.3%) | 9 (1.4%) |
| Stage 3 | 14 (16%) | 679 (21%) | | 7 (16%) | 551 (22%) | 3 (12%) | 14 (16%) | 4 (21%) | 114 (18%) |
| Stage 4 | 70 (80%) | 2,167 (68%) | | 36 (82%) | 1,624 (66%) | 20 (80%) | 65 (76%) | 14 (74%) | 478 (75%) |
| Unknown | 50 | 2,214 | | 29 | 1,482 | 9 | 54 | 12 | 678 |
| **DNA assay** | | | 0.3 | | | | | | |
| xE | 0 (0%) | 69 (1.3%) | | 0 (0%) | 66 (1.7%) | 0 (0%) | 3 (2.2%) | | |
| xE.v2 | 2 (1.4%) | 88 (1.6%) | | 1 (1.4%) | 45 (1.1%) | 0 (0%) | 1 (0.7%) | 1 (3.2%) | 42 (3.2%) |
| xF | 0 (0%) | 1 (<0.1%) | | 0 (0%) | 1 (<0.1%) | | | | |
| xO | 2 (1.4%) | 62 (1.1%) | | 1 (1.4%) | 62 (1.6%) | 1 (2.9%) | 0 (0%) | | |
| xT | 4 (2.9%) | 80 (1.5%) | | 4 (5.5%) | 80 (2.0%) | | | | |
| xT.v2 | 6 (4.3%) | 403 (7.5%) | | 4 (5.5%) | 386 (9.8%) | 2 (5.9%) | 17 (12%) | | |
| xT.v3 | 14 (10%) | 401 (7.4%) | | 9 (12%) | 378 (9.6%) | 5 (15%) | 23 (17%) | | |
| xT.v4 | 110 (80%) | 4,296 (80%) | | 54 (74%) | 2,924 (74%) | 26 (76%) | 95 (68%) | 30 (97%) | 1,277 (97%) |
| **Sequencing year** | | | 0.4 | | | | | | |
| 2017-2019 | 16 (12%) | 665 (12%) | | 12 (16%) | 639 (16%) | 4 (12%) | 26 (19%) | 0 (0%) | 0 (0%) |
| 2020 | 35 (25%) | 1,052 (19%) | | 21 (29%) | 1,001 (25%) | 14 (41%) | 51 (37%) | 0 (0%) | 0 (0%) |
| 2021 | 29 (21%) | 1,425 (26%) | | 18 (25%) | 1,373 (35%) | 11 (32%) | 52 (37%) | 0 (0%) | 0 (0%) |
| 2022 | 52 (38%) | 2,040 (38%) | | 22 (30%) | 929 (24%) | 5 (15%) | 10 (7.2%) | 25 (81%) | 1,101 (83%) |
| 2023 | 6 (4.3%) | 218 (4.0%) | | 0 (0%) | 0 (0%) | 0 (0%) | 0 (0%) | 6 (19%) | 218 (17%) |
| **Scanner Make** | | | <0.001 | | | | | | |
| Aperio | 118 (86%) | 3,909 (72%) | | 65 (89%) | 2,873 (73%) | 33 (97%) | 110 (79%) | 20 (65%) | 926 (70%) |
| Philips | 20 (14%) | 1,491 (28%) | | 8 (11%) | 1,069 (27%) | 1 (2.9%) | 29 (21%) | 11 (35%) | 393 (30%) |
| **Tumor mutational burden** | | | <0.001 | | | | | | |
| High | 125 (93%) | 47 (0.9%) | | 66 (93%) | 39 (1.0%) | 32 (97%) | 2 (1.5%) | 27 (90%) | 6 (0.5%) |
| Low | 9 (6.7%) | 5,199 (99%) | | 5 (7.0%) | 3,763 (99%) | 1 (3.0%) | 131 (98%) | 3 (10%) | 1,305 (100%) |
| Unknown | 4 | 154 | | 2 | 140 | 1 | 6 | 1 | 8 |
| **Sample collection year** | | | | | | | | | |
| Before 2000 | 0 (0%) | 5 (<0.1%) | | 0 (0%) | 3 (<0.1%) | 0 (0%) | 0 (0%) | 0 (0%) | 2 (0.2%) |
| 2000-2009 | 0 (0%) | 95 (1.8%) | | 0 (0%) | 77 (2.0%) | 0 (0%) | 3 (2.2%) | 0 (0%) | 15 (1.1%) |
| 2010-2014 | 4 (2.9%) | 391 (7.3%) | | 3 (4.2%) | 326 (8.4%) | 0 (0%) | 8 (5.8%) | 1 (3.2%) | 57 (4.3%) |
| 2015-2019 | 48 (35%) | 1,938 (36%) | | 29 (40%) | 1,648 (42%) | 15 (44%) | 61 (44%) | 4 (13%) | 229 (17%) |
| 2020 | 27 (20%) | 804 (15%) | | 16 (22%) | 695 (18%) | 9 (26%) | 33 (24%) | 2 (6.5%) | 76 (5.8%) |
| 2021 | 24 (18%) | 1,067 (20%) | | 13 (18%) | 868 (22%) | 8 (24%) | 33 (24%) | 3 (9.7%) | 166 (13%) |
| 2022 | 33 (24%) | 1,044 (19%) | | 11 (15%) | 280 (7.2%) | 2 (5.9%) | 1 (0.7%) | 20 (65%) | 763 (58%) |
| 2023 | 1 (0.7%) | 11 (0.2%) | | 0 (0%) | 0 (0%) | 0 (0%) | 0 (0%) | 1 (3.2%) | 11 (0.8%) |
| Unknown | 1 | 45 | | 1 | 45 | | | | |
| **ADT prior to collection** | 21 (21%) | 448 (14%) | 0.043 | 13 (25%) | 356 (14%) | 6 (23%) | 12 (14%) | 2 (10%) | 80 (13%) |
| Unknown | 39 | 2,203 | | 20 | 1,457 | 8 | 53 | 11 | 693 |

[1] n (%)
[2] Fisher's exact test; Pearson's Chi-squared test

Supplementary Table 2. Multivariate logistic regression model for clinical and molecular variables in patient cohort.

```
                  Generalized Linear Model Regression Results
==============================================================================
Dep. Variable:             msi_status   No. Observations:                 3045
Model:                            GLM   Df Residuals:                     3032
Model Family:                Binomial   Df Model:                           12
Link Function:                  Logit   Scale:                          1.0000
Method:                          IRLS   Log-Likelihood:                -113.15
Date:                Mon, 10 Jul 2023   Deviance:                       226.30
Time:                        22:03:32   Pearson chi2:                 2.11e+05
No. Iterations:                     9   Pseudo R-squ. (CS):             0.1804
Covariance Type:            nonrobust
==============================================================================
                                coef    std err          z      P>|z|      [0.025      0.975]
------------------------------------------------------------------------------
Intercept                   -18.6076      3.554     -5.236      0.000     -25.573     -11.643
age_days                   5.017e-05   6.32e-05      0.794      0.427   -7.37e-05       0.000
race_concept_canonical_name   0.2563      0.251      1.021      0.307      -0.236       0.748
histology_rollup              0.5161      0.696      0.742      0.458      -0.847       1.880
total_gleason                 0.5232      0.252      2.075      0.038       0.029       1.018
stage                         0.0622      0.132      0.472      0.637      -0.196       0.321
dna_sequencing_date_days      0.0002      0.001      0.276      0.783      -0.001       0.001
collection_date_days          0.0007      0.000      2.446      0.014       0.000       0.001
procedure_type                0.3725      0.226      1.646      0.100      -0.071       0.816
scanner_make                 -0.7097      0.487     -1.457      0.145      -1.664       0.245
dna_assay                     0.3281      0.196      1.674      0.094      -0.056       0.712
tmb_status                    7.0011      0.457     15.333      0.000       6.106       7.896
adt_status                   -0.0874      0.738     -0.118      0.906      -1.533       1.359
==============================================================================
```

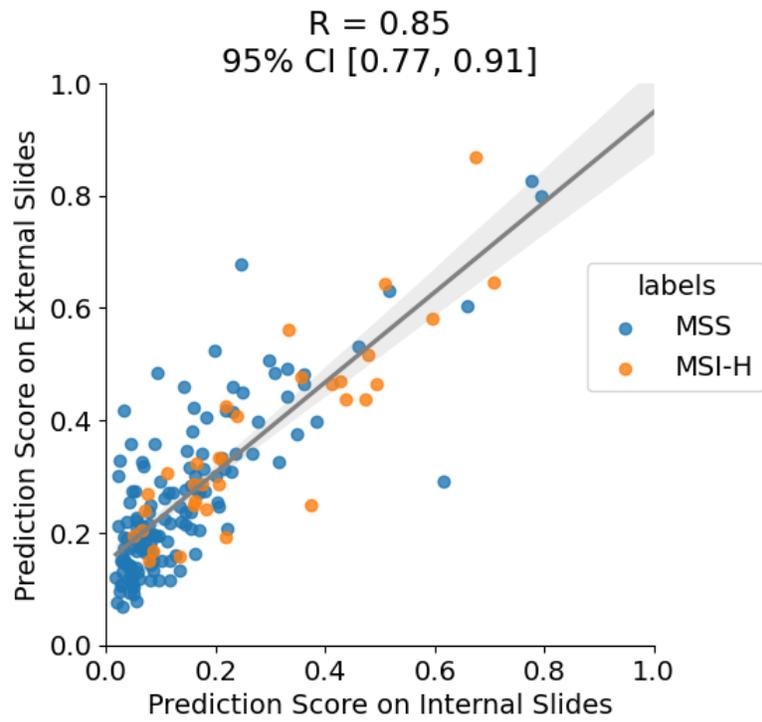

Supplementary Figure 1. Correlation between the prediction scores of the internally and externally stained and scanned slides in the paired validation set.

(a)

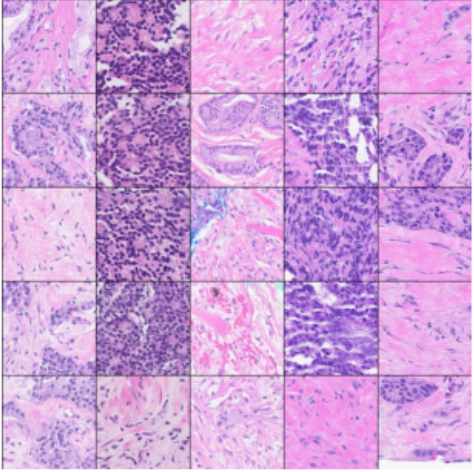
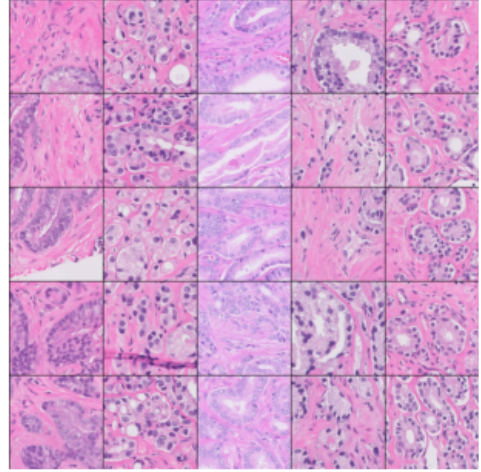
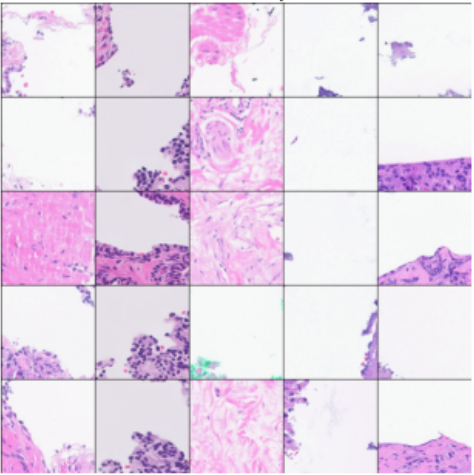
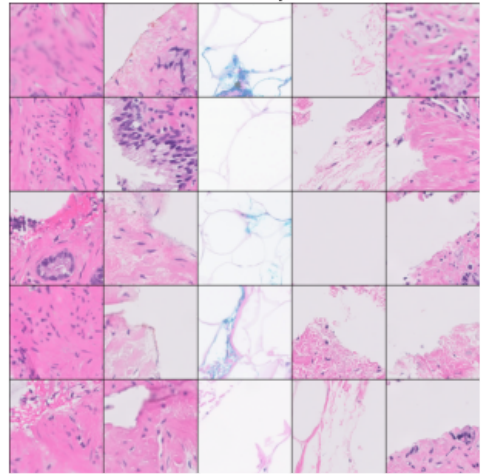

(b)

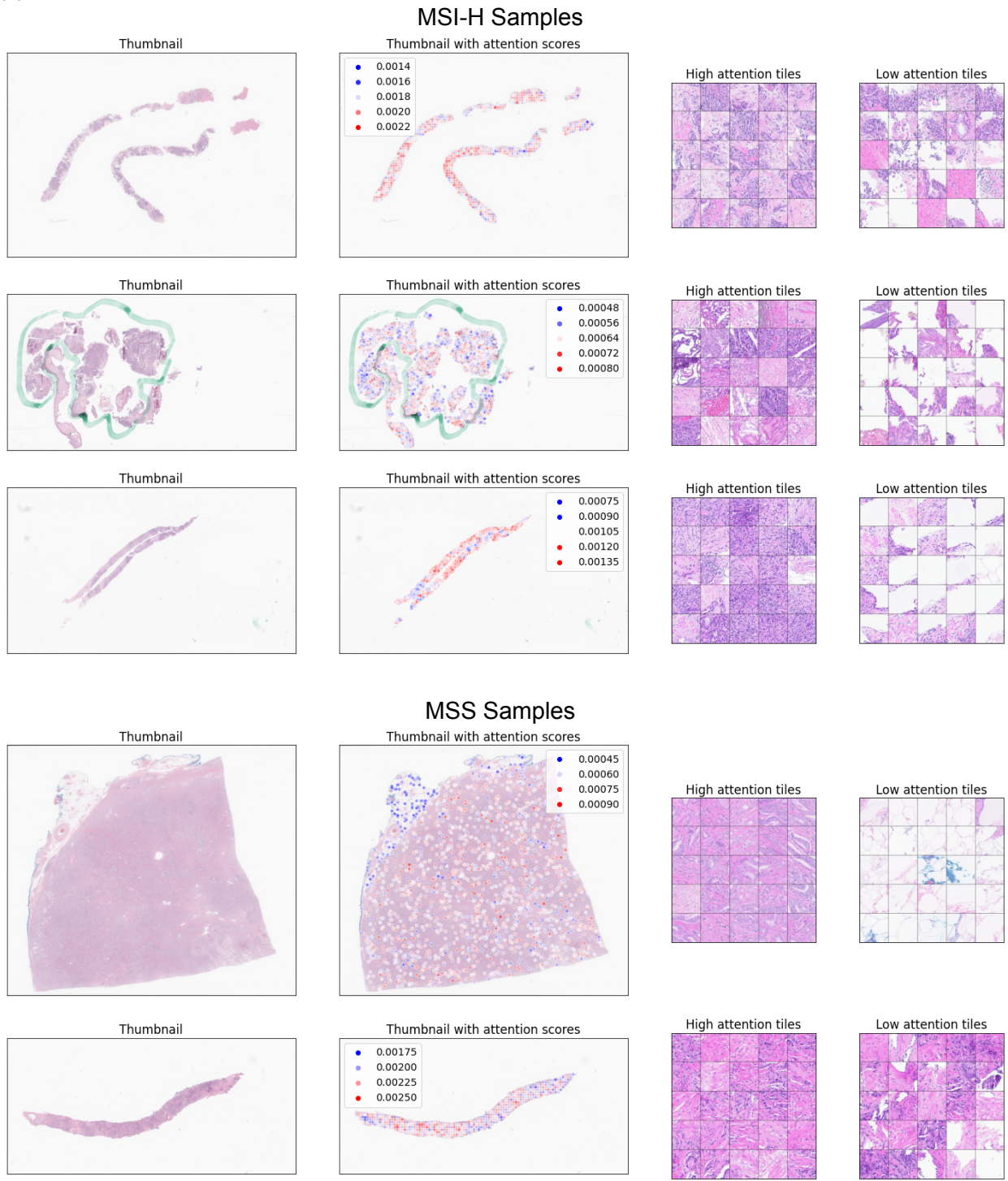

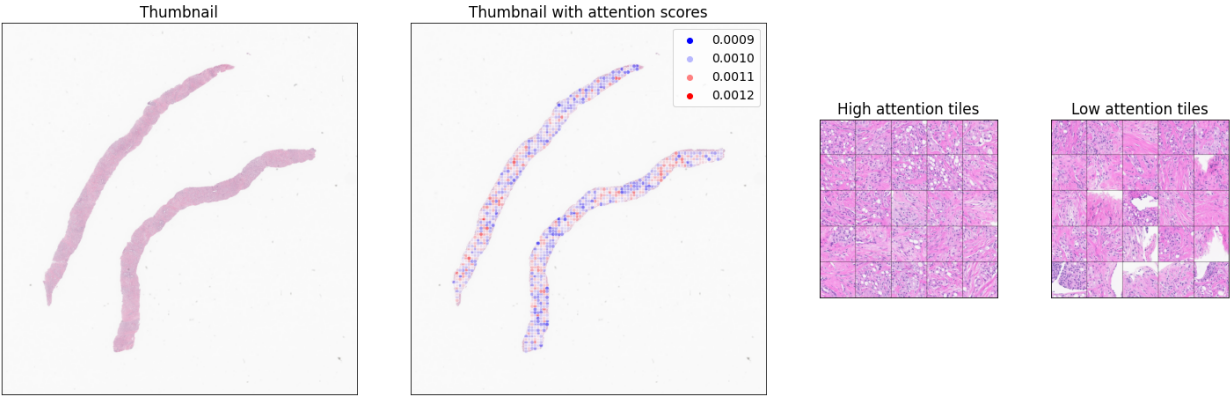

Supplementary Figure 2. (a) High-attention tiles (top) and low-attention tiles (bottom) from WSIs that received the highest prediction scores in their corresponding correct class. (b) Examples of MSI-H (top) and MSS (bottom) slides showing the thumbnail of each WSI, attention score heatmaps overlaid on the WSI, and top high/low-attention tiles from each slide.

(a) Paired validation set (internal scans)

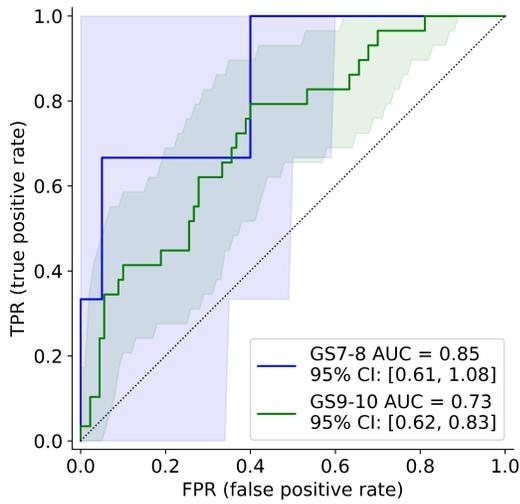
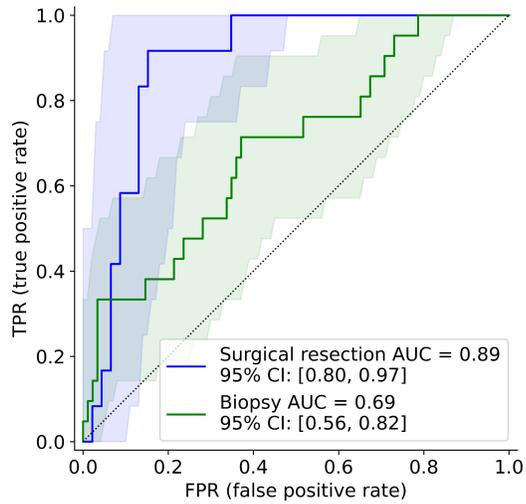
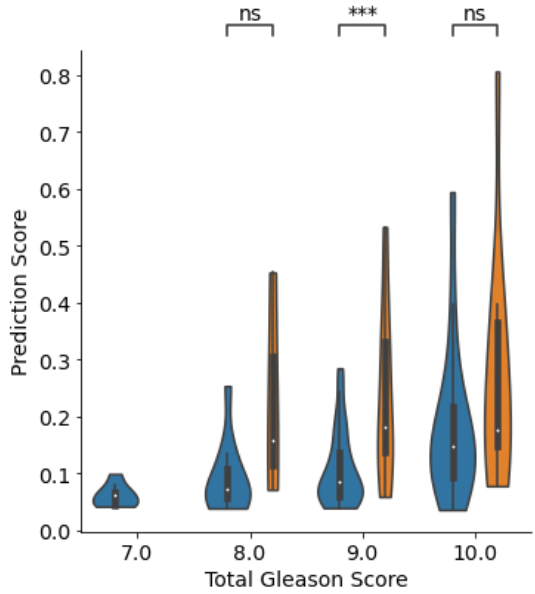
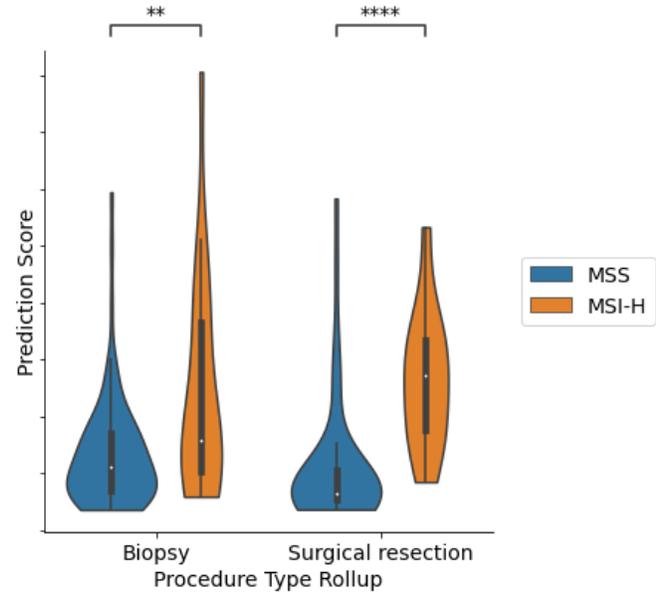

(b) Paired validation set (external scans)

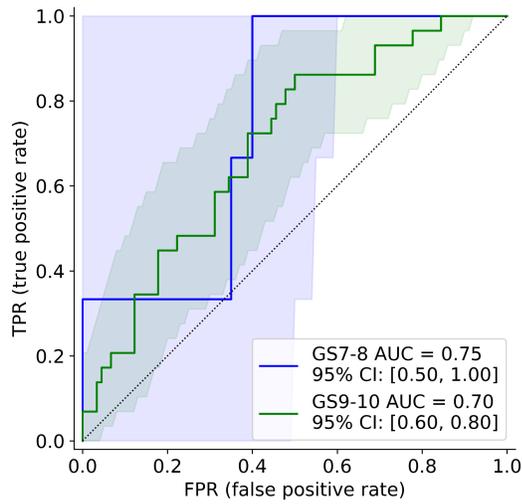
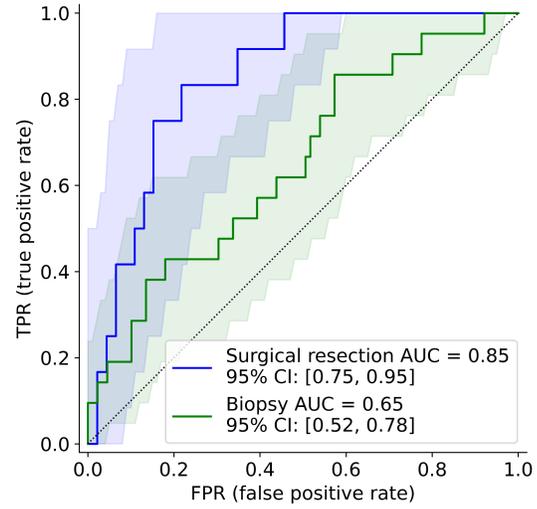
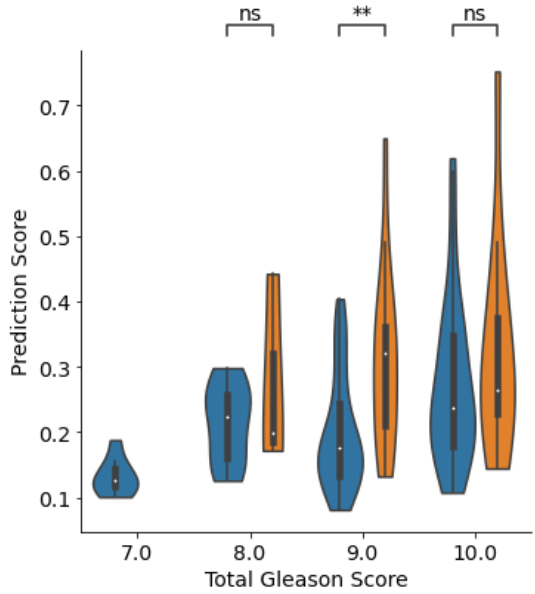
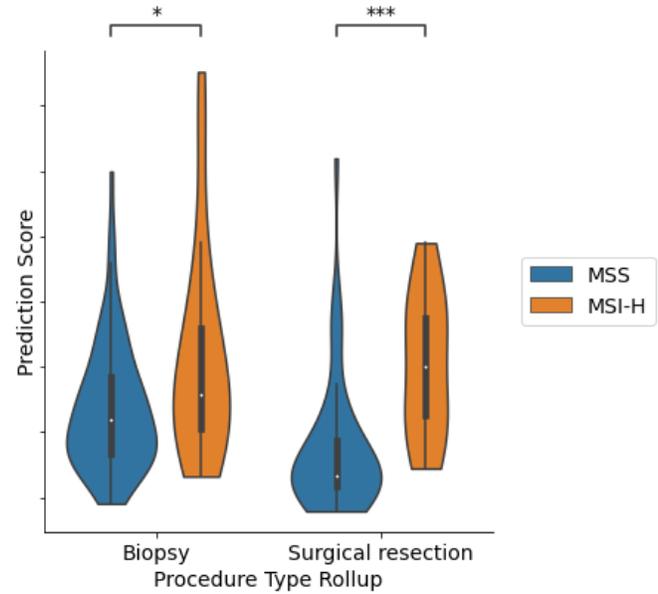

(c) Temporal validation set

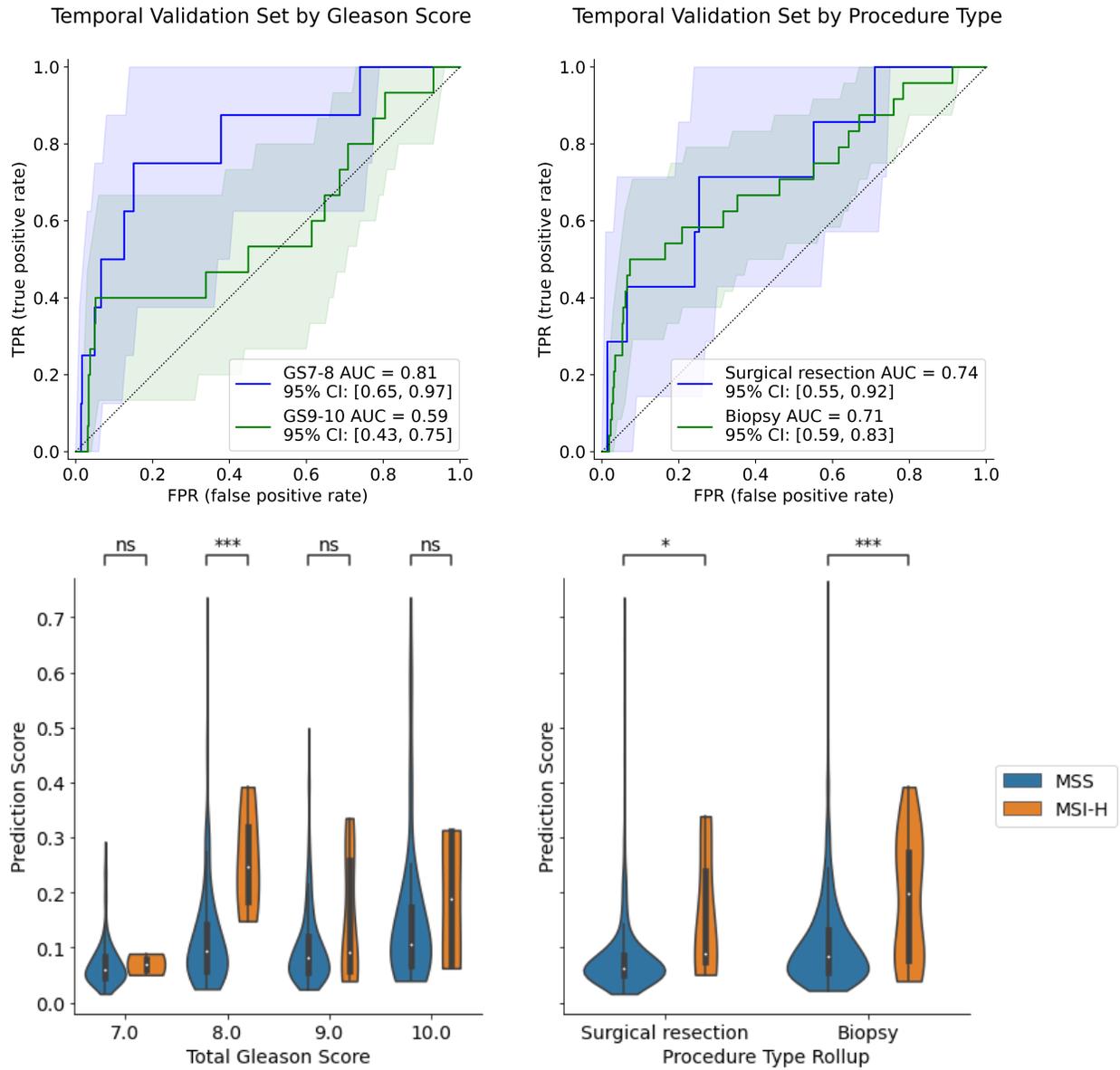

Supplementary Figure 3. Receiver operating characteristic (ROC) curves and violin plots of prediction scores for various clinical subgroups in (a) internal scans in the paired validation set, (b) external scans in the paired validation set, and (c) temporal validation set. P-value annotation legend: ns: $p \leq 1$, *: $0.01 < p \leq 0.05$, **: $0.001 < p \leq 0.01$, ***: $0.0001 < p \leq 0.001$, ****: $p \leq 0.0001$.

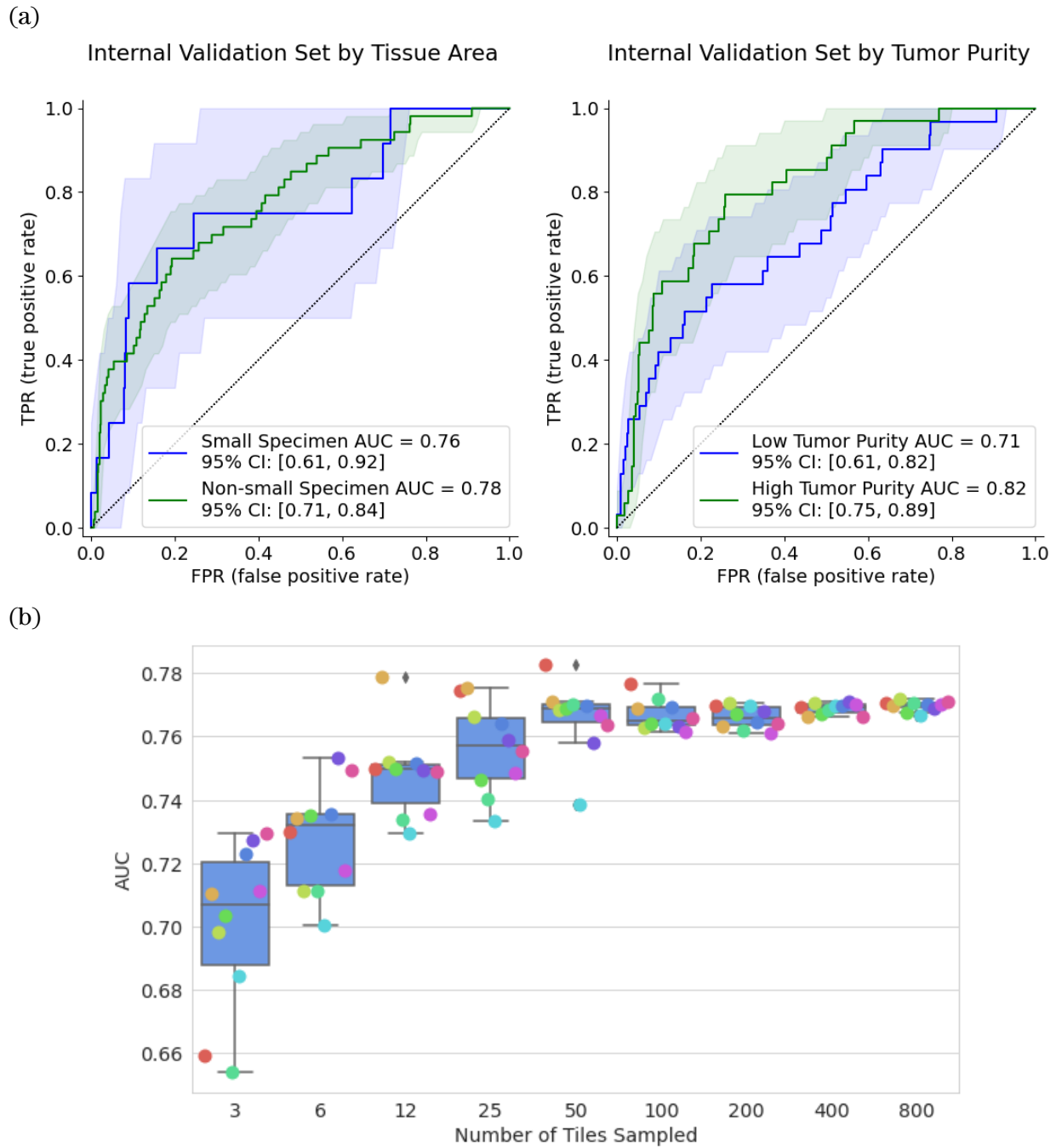

Supplementary Figure 4. (a) Receiver operating characteristic (ROC) curves for tissue area and tumor purity subgroups in the internal scans in the pooled validation set that combines the internal scans of the paired validation set and the temporal validation set. The threshold value for both subgroups is the first quartile within the validation set, which is 9.35 mm$^2$ for tissue area and 50% for tumor purity. (b) Effect of tissue area on model performance from the simulation experiment on the combined internal test set. Data points within each box demonstrate variations when 10 random different seeds were used for tile sampling.

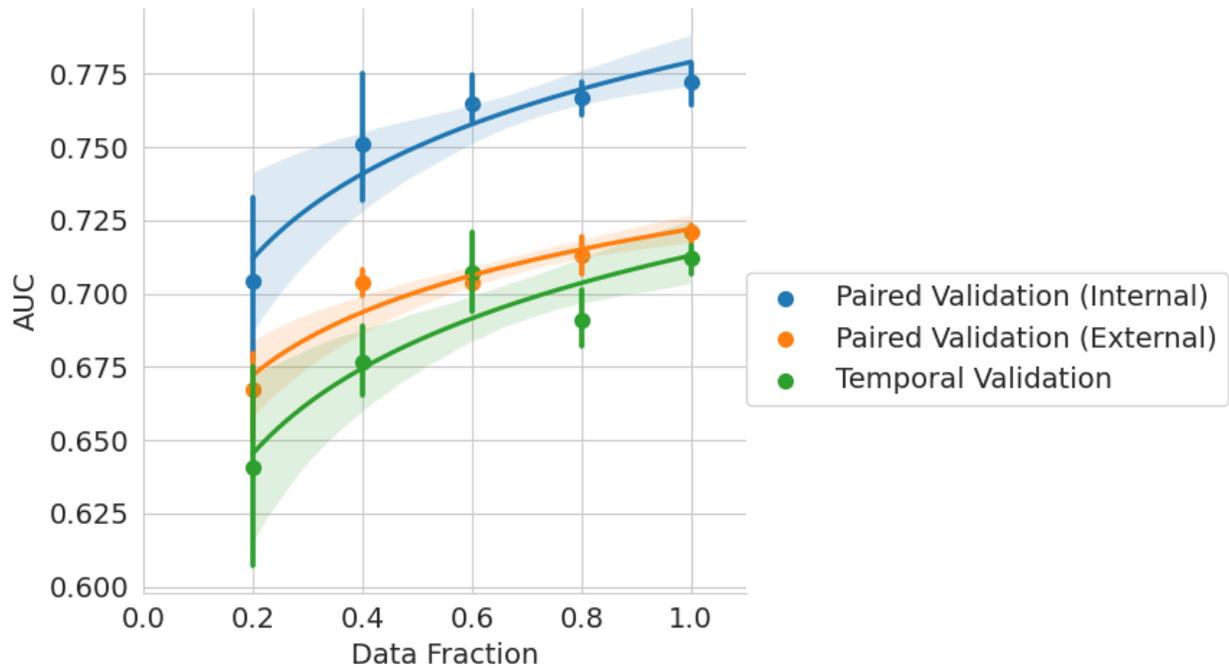

Supplementary Figure 5. Area under the receiver operating characteristic curves (AUC) trends on the three validation sets when various fractions of the training data were used. The lines are fitted linear regressions with the explanatory variable on the log scale.

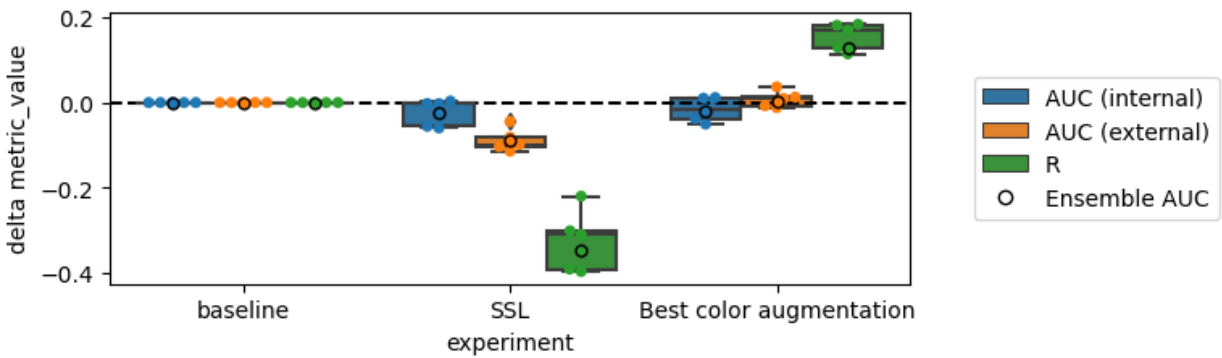

Supplementary Figure 6. The difference in metrics caused by self-supervised learning pretraining and strong slide-level color augmentation. The metrics include the area under the receiver operating characteristic curves (AUC) on the internally and externally stained and scanned slides in the paired validation set and the Pearson correlation, R, between prediction scores between the internally and externally stained and scanned slides in the paired validation set. The baseline model was initiated with ImageNet weights and used tile-level color augmentation with a set of default hyperparameters.